\documentclass[conference]{IEEEtran}  

\IEEEoverridecommandlockouts                              

\makeatletter

\let\proof\@undefined
\let\endproof\@undefined
\makeatother

\usepackage{graphicx} 
\usepackage{url}
\usepackage{epstopdf}
\DeclareGraphicsExtensions{.eps}

\usepackage{multirow}
\usepackage{algorithmic}
\usepackage{algorithm}
\usepackage{amsmath} 
\usepackage{amssymb}
\usepackage{amsxtra}
\usepackage{amsfonts}
\usepackage{color} 
\usepackage{bm}
\usepackage{placeins}
\usepackage[caption=false]{subfig}
\usepackage{xspace}
\usepackage{rotating}
\usepackage{siunitx}
\usepackage{textcomp}%
\usepackage{algorithm}
\usepackage{algorithmic}

\newcommand{\myvector}[1]{\bm{#1}}
\newcommand{\myvec}[1]{\myvector{#1}}

\newcommand{\R}[1]{\mathbb{R}^{#1}}

\newcommand{\argmax}{\operatornamewithlimits{arg max}}

\newcommand{\algorithmicinput}{\textbf{Input:}}
\newcommand{\algorithmicoutput}{\textbf{Output:}}
\newcommand{\INPUT}{\item[\algorithmicinput]}
\newcommand{\OUTPUT}{\item[\algorithmicoutput]}

\newcommand{\dist}[1]{\SI{#1}{\meter}}

\newcommand{\dgr}[1]{\ensuremath{#1^{\circ}}}

\newcommand{\gauss}[2]{\mathcal{N}(#1 , #2)}

\newcommand{\excise}[1]{}

\newcommand{\X}{S}
\newcommand{\x}{\ensuremath{\myvec{s}}}

\newcommand{\U}{A}
\newcommand{\uv}{\ensuremath{\myvec{a}}}
\newcommand{\ui}[1]{a}

\newcommand{\g}{\ensuremath{\myvec{g}}}

\newcommand{\ac}{\uv}

\title{
PRM-RL: Long-range Robotic Navigation Tasks by Combining Reinforcement Learning and Sampling-based Planning\\[1.75ex] 
{\normalfont\large 
Aleksandra Faust$^1$, Oscar Ramirez$^1$, Marek Fiser$^1$, Kenneth Oslund$^1$,\\ Anthony Francis$^1$, James Davidson$^1$, and Lydia Tapia$^2$%
}\\
}

\newcommand\Mark[1]{\textsuperscript{#1}}

\author{
	\IEEEauthorblockA{%
		\Mark{1}Google Brain\\
		Mountain View, CA, USA%
	}
	\and
	\IEEEauthorblockA{%
		\Mark{2}Department of Computer Science\\
		University of New Mexico, Albuquerque, NM, USA%
	}
}

\begin{document}

\maketitle
\thispagestyle{empty}
\pagestyle{empty} 

\begin{abstract}
We present PRM-RL, a hierarchical method for long-range navigation task completion that combines sampling-based path planning with reinforcement learning (RL).
The RL agents learn short-range, point-to-point navigation policies that capture robot dynamics and task constraints without knowledge of the large-scale topology. Next, the sampling-based planners provide roadmaps which connect robot configurations that can be successfully navigated by the RL agent. The same RL agents are used to control the robot under the direction of the planning, enabling long-range navigation. We use the Probabilistic Roadmaps (PRMs) for the sampling-based planner. The RL agents are constructed using feature-based and deep neural net policies in continuous state and action spaces. We evaluate PRM-RL, both in simulation and on-robot, on two navigation tasks with non-trivial robot dynamics: end-to-end differential drive indoor navigation in office environments, and aerial cargo delivery in urban environments with load displacement constraints. Our results show improvement in task completion over both RL agents on their own and traditional sampling-based planners. In the indoor navigation task, PRM-RL successfully completes up to \dist{215} long trajectories under noisy sensor conditions, and the aerial cargo delivery completes flights over \dist{1000} without violating the task constraints in an environment 63 million times larger than used in training.
\end{abstract}  

\section{Introduction}
\label{sec:1}

\begin{figure}[h!]
\label{fig:tasks}
	\begin{center}
		\begin{tabular}{cc}
			\subfloat[Indoor navigation]{\includegraphics[trim=0mm 0mm 0mm 0mm,clip,width=0.25\textwidth,height=5.25cm,keepaspectratio=true]{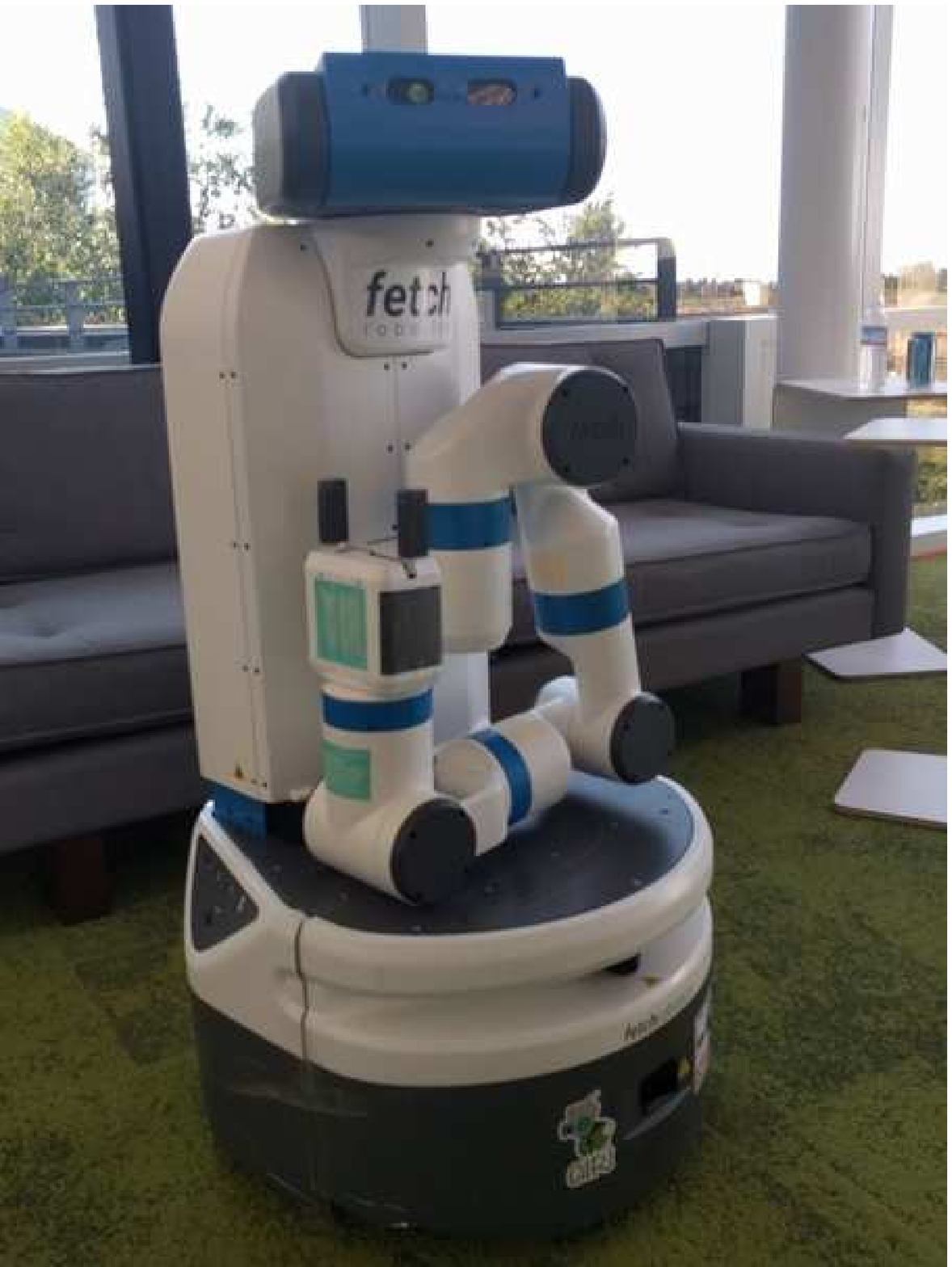}\label{fig:fetch}}
			&
			\subfloat[Aerial cargo delivery]{\includegraphics[trim=0mm 0mm 0mm 0mm,clip,width=0.2\textwidth,height=5.25cm,keepaspectratio=false]{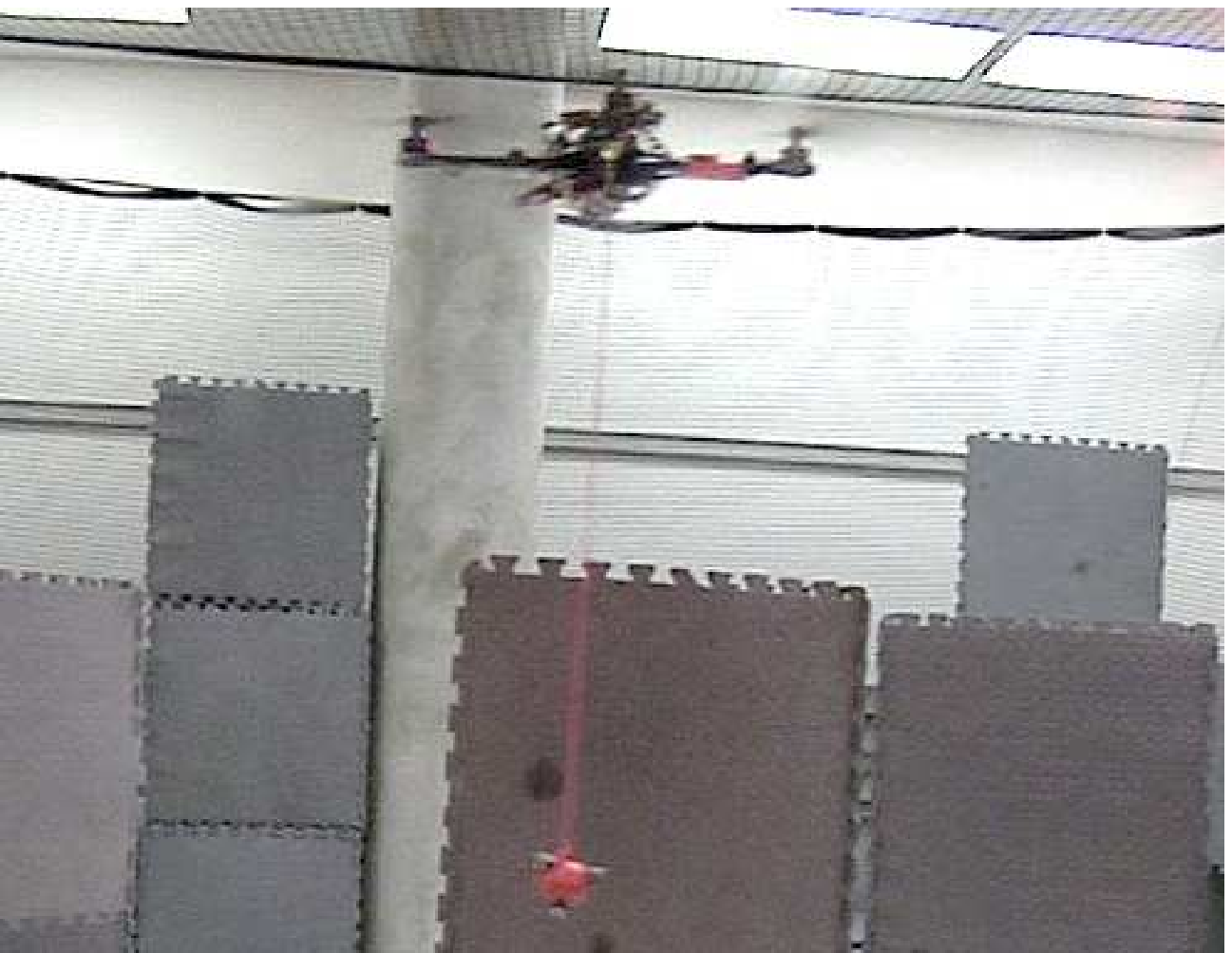}\label{fig:quad}}
			\\
		\end{tabular}
		\caption{\small Case studies of long-range navigation tasks.}
	\end{center}
\end{figure}

Long-range navigation tasks require robots to move safely over substantial distances while satisfying task constraints.
For example, indoor navigation (Fig. \ref{fig:fetch}) requires a robot to navigate through buildings avoiding obstacles using noisy sensor data.
As another example, an aerial cargo delivery \cite{faust-ai-journal} requires a suspended load-equipped unmanned aerial vehicle (UAV) to fly over long distances while avoiding obstacles and minimizing the oscillations of the load (Fig. \ref{fig:quad}).
We factor the long-range navigation task into two parts: long-range collision-free path finding and local robot control. Collision-free path finding identifies an obstacle-free path from a starting position to a distant goal while avoiding obstacles \cite{Lavalle06book}.
Local robot control produces feasible controls that the robot executes to perform the task while both satisfying task constraints and staying near the obstacle-free path.

Sampling-based planners, such as Probabilistic Roadmaps (PRMs) \cite{kavraki-prm} and Rapidly Exploring Random Trees (RRTs) \cite{kuffner-rrt,Lavalle-rrt2}, efficiently solve this problem by approximating the topology of the configuration space (C-space), the space of all possible robot configurations.
These methods construct a graph or tree by sampling points in C-space, and connecting points if there is a collision-free local path between points. Typically this local path is created by a line of sight test or an inexpensive local planner.

Regardless of how a collision-free path is generated, executing it introduces new complications.
A robot cannot simply follow the C-space path, but must 1) satisfy the constraints of the task, 2) handle changes in the environment, and 3) compensate for sensor noise, measurement errors, and unmodeled system dynamics.
Reinforcement learning (RL) agents have emerged as a viable approach to robot control \cite{jan-peters-ijrr-survey}.
RL agents have solved complex robot control problems \cite{DBLP:journals/corr/YahyaLKCL16}, adapted to new environments \cite{faust-ai-journal}, demonstrated robustness to noise and errors \cite{faust-icra-15}, and even learned complex skills \cite{bartt-mkp-11}; however, RL agents can be hard to train if rewards are sparse \cite{pearl}.
This presents both opportunities and challenges in applying RL to navigation. 
RL agents have been successfully applied to several permutations of the navigation task and video games \cite{atari-paper}, making them good choices to deal with task constraints.
Conversely, long-range navigation over complex maps has sparse rewards, making agents either difficult to train or successful only at short range because of vulnerabilities to local minima. For example, both city maps and office floorplans frequently have goals on the other side of wide barriers, which can cause local agents to become confused, or on the other side of box canyons, where local agents can become trapped.

\begin{figure*}[t]
	\begin{center}
		\begin{tabular}{cc}
			\subfloat[Training environment - \dist{23} by \dist{18}]{\includegraphics[trim=0mm 0mm 0mm 0mm,clip,width=0.5\textwidth,height=4.0cm,keepaspectratio=true]{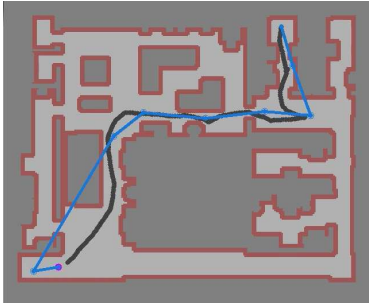}\label{fig:mtv1965}} 
			&
			\subfloat[Building 1 - \dist{183} by \dist{66}]{\includegraphics[trim=0mm 0mm 0mm 0mm,clip,width=0.5\textwidth,height=4.0cm,keepaspectratio=true]{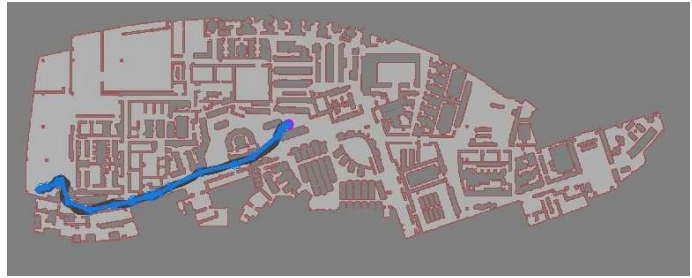}\label{fig:b40}}\\
			\subfloat[Building 2 - \dist{60} by \dist{47}]{\includegraphics[trim=0mm 0mm 0mm
			0mm,clip,width=0.5\textwidth,height=5.25cm,keepaspectratio=true]{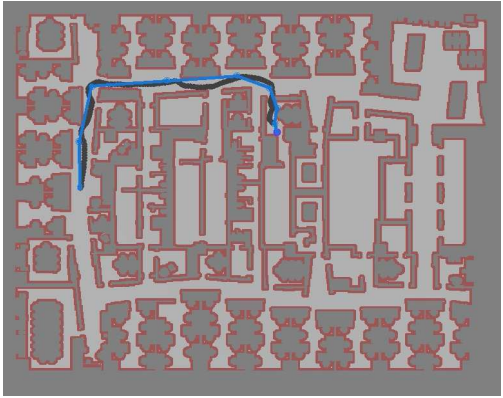}\label{fig:sfo}} 
			&
			\subfloat[Building 3 - \dist{134} by \dist{93} ]{\includegraphics[trim=0mm 0mm 0mm 0mm,clip,width=0.5\textwidth,height=5.25cm,keepaspectratio=true]{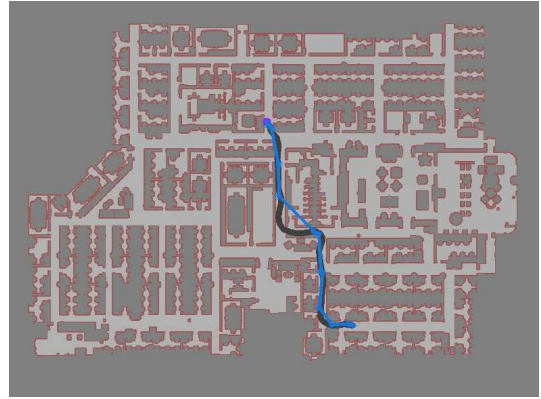}\label{fig:1667}}\\
		\end{tabular}
		\caption{\small The environments used for the indoor navigation tasks are derived from real building plans. a) The smallest environment is used to train the RL agent. b)-d) The PRMs are built using agents trained in the training environment. Red regions are deemed too close to obstacles and cause episode termination when the robot enters them; white is free space from where the start and goals are selected. Line-connected PRM waypoints (blue) and agent executed trajectory with RL agent (black).\label{fig:maps}}
	\end{center}
\end{figure*}

We present PRM-RL, an approach to long-range navigation tasks which overcomes the limitations of PRMs and RL agents by using them to address each other's shortfalls.
In PRM-RL, an RL agent is trained to execute a point-to-point task locally, learning the task constraints, system dynamics and sensor noise independent of the long-range environment structure.
Then, PRM-RL builds a roadmap using this RL agent to determine connectivity, rather than the traditional collision-free straight-line interpolation in C-space.
PRM-RL connects two configuration points only if the RL agent can consistently perform the local point-to-point task between them and all configurations along the produced trajectory are collision-free.
The roadmap thus learns the long-range environment structure that can be navigated using that RL agent.
Compared to roadmaps constructed based on pure C-space linear connectivity, PRM-RL roadmaps can obey robot dynamics and task constraints.
The roadmap also learns to avoid local minima that cause failures of the RL agent.
The resulting long-range navigation planner thus combines the planning efficiency of a PRM with the robustness of an RL agent, while avoiding local-minima traps, and providing resilience in the face of moderate changes to the environment. 

\begin{figure}[]
\centering
\includegraphics[width=0.48\textwidth,height=1.5cm,keepaspectratio=false]{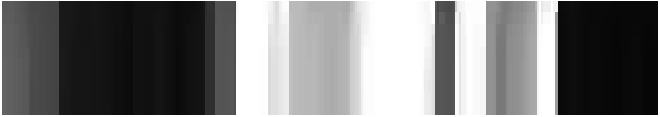}
\caption{\small Lidar observation that the differential drive robot uses for navigation. This observation corresponds to a hallways with a clear path ahead, and walls to the left and to the right. The white rays mean there are not objects within \dist{5}, and black rays mean that there is an obstacle near. Notice the sensor noise in the center.\label{fig:lidar}}
\end{figure}

To evaluate the approach, we focus on two problems: indoor navigation (Fig. \ref{fig:fetch}), and aerial cargo-delivery (Fig. \ref{fig:quad}).
The indoor navigation problem requires a differential drive robot to navigate inside buildings (Fig. \ref{fig:maps}) while avoiding  static obstacles using only its LIDAR sensor (Fig. \ref{fig:lidar}).
We build PRMs from blueprints of target buildings, training RL agents on the smallest map using a simulator with noisy sensors and dynamics designed to emulate the unprocessed, noisy sensor input of the actual robot.
We evaluate planning and execution of the PRMs in environments that the agent was not trained in. 
The aerial cargo delivery problem requires a quadrotor UAV with a suspended load to transport the cargo with minimum residual oscillations while maintaining load displacement below a given upper bound. 
The combined quadrotor-load system is non-linear and unstable.
We evaluate the planner in a simulated urban environment and in two experimental environments, assessing adherence to task and dynamic constraints. 
We show that in environments with static obstacles, the planner maintains task constraints over long trajectories (over 1 km in length).
Finally, results on physical robots for both problems show the PRM-RL approach maintains system dynamic constraints while producing feasible trajectories.

\section{Related Work}
\label{sec:2}

PRMs have been used in a wide variety of planning problems from robotics \cite{latombe-lunar,malone-embodied2012} to molecular folding \cite{AlBluwi_survey_2012,park-prmrl,tapia_motion_2010}.
They have also been integrated with reinforcement learning for state space dimensionality reduction \cite{malone-journal-14,park-prmrl} by using PRM nodes as state space for the reinforcement learning agent. 
In contrast, our work applies reinforcement learning on the full state space as a local planner for PRMs. 
In prior work, for an aerial cargo delivery task, we trained RL agents to track paths generated from PRMs constructed using a straight line local planner \cite{faust-ai-journal}.
In this work, the RL agents themselves act as the local planner, eliminating the need to train separate tracking agents.
PRMs have also been modified to work with moving obstacles \cite{hsu-moving-02, amato-moving-07}, noisy sensors  \cite{malone-iros-2013}, and localization errors \cite{amato-uncertain-13, alterovitz-rss-07}.
Safety PRM \cite{malone-iros-2013} uses probabilistic collision checking with straight-line planner, associating with all nodes and edges a measure of potential collision.
In our method, the RL local planner does Monte Carlo path rollouts with deterministic collision checking but noisy sensors and dynamics.
We only add edges if the path can be consistently navigated.

Reinforcement learning has recently gained popularity in solving motion planning problems for systems with unknown dynamics \cite{jan-peters-ijrr-survey}, and has enabled robots to learn tasks that have been previously difficult or impossible \cite{abel2016exploratory,chen2015deepdriving,levine2016end}. 
Deep Deterministic Policy Gradient (DDPG) \cite{ddpg} is a current state-of-the-art algorithm that works with very high dimensional state and action spaces and is able to learn to control robots based on unprocessed sensor observations \cite{levine2016end}. 
Continuous Action Fitted Value Iteration (CAFVI) \cite{faust-acta-13} is a feature-based continuous state and action reinforcement algorithm that has been used for problems such as multi-robot tasks \cite{faust-acta-13}, flying inverted pendulums \cite{figueroa-wcica-14}, and obstacle avoidance \cite{faust-icra-15}.
In this work, we use DDPG as the local planner for the indoor navigation task, and CAFVI as the planner for the aerial cargo delivery task. 
Model-predictive control (MPC) \cite{nmpc-2011}, action filtering \cite{faust-ai-journal}, and hierarchical policy approximations \cite{mansley-hoot-11}, like RL, provide policies which respect robot dynamics and task constraints. However, they are computationally more expensive than RL at the execution time, making them not practical for building computationally-demanding roadmaps.

\section{Methods}
\label{sec:3}
\begin{figure}[hbtp]
\centering
\includegraphics[scale=0.50]{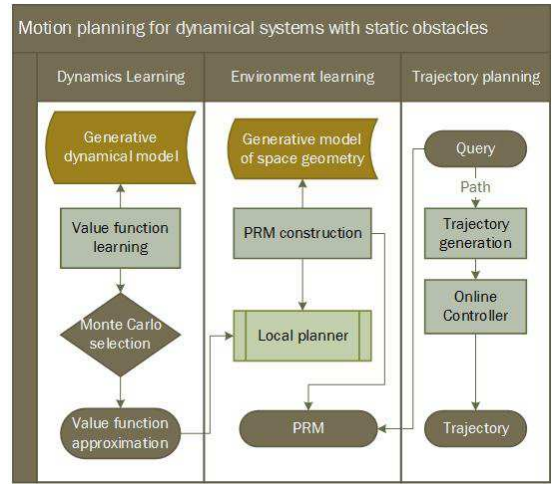}
\caption{PRM-RL flowchart.}
\label{fig:arch}
\end{figure}

PRM-RL works in three stages: RL agent training, roadmap creation, and roadmap querying. 
Fig. \ref{fig:arch} shows the overview of the method.
A key feature which makes PRM-RL transferable to new environments is that it learns task and system dynamics separately from the deployment environment.
In the first stage, we train an RL agent to perform a task on an environment comparable to the deployment environment, but with a small state space to make learning more tractable.
The RL agent training stage is a Monte Carlo simulation process: regardless of the learning  algorithm used (DDPG, CAFVI or another), we train multiple policies and select the fittest one for the next stage of PRM-RL.
This best policy, or value function, is then passed to the roadmap creation stage. 
The PRM builder learns a roadmap for a particular deployment environment using the best RL agent as a local planner using Algorithm \ref{alg:add_edge}. 
Constructed roadmap can be used for number of queries in the environment, as long as the same RL agent is used for execution.
While the RL agent is robot/task dependent, it is environment independent, and it can be used to build roadmaps for a multitude of environments, as we show in the results. 

\subsection{RL agent training}

PRM-RL takes a start state $\myvec{s}$ and a goal state $\myvec{g}$ as two valid points in the robot's state space $\X$. 
The robot state space $\X$ is set of all possible robot observations in the control space for the robot, and is thus a superset of the configuration space, C-space. 
A state space point $\x \in \X$ is valid if and only if it satisfies the task constraints for some predicate $L(\x)$, and the point's projection onto C-space $p(\x)$ belongs in C-free, a partition of C-space consisting of only collision-free points. 
When either of the conditions is violated, the system's safety cannot be guaranteed, and the task cannot be performed. 
The task is completed when the system is sufficiently close to the goal state, $\|p(\myvec{s})-p(\myvec{g})\| \leq \epsilon$ in the configuration space.
Our goal is to find a transfer function 
\begin{equation}
\label{eq:dynSystem}
\dot{\x} = f(\x, \uv).
\end{equation}
that leads the system to task completion. 
Formally, in the Markov Decision Process (MDP) setting, our goal is to find a policy $\pi:\X \rightarrow \U$, such that for an initial state $\myvec{s}$, there is $n_0 > 0$, such that $\forall n \leq n_0\, \x_n=\pi^{n}(\myvec{s})$ and $L(\x_n)$ holds, and $\|p(\x_{n_0}) - p(\myvec{g})\| \leq \epsilon$. 

The RL agent is trained to perform point to point navigation task without memory of the workspace topology. Reinforcement learning finds a solution to a discrete time MDP with continuous multi-dimensional state and action spaces, given as a tuple of states, action, transitions, and rewards, $(\X, \U, P, R)$. $\X \subset \R{d_s}$ is the state or observation space of the robot. For the indoor navigation task, we use the combined space of the robot's position relative to the goal in polar coordinates, $\myvec{g}$, and all possible LIDAR observations (Fig. \ref{fig:lidar}), $\myvec{o}$, casting 64 rays over \dgr{220} field-of-view with up to \dist{5} depth. Overall, the state is $\x = (\myvec{g}, \myvec{o}) \in \R{66}.$ In the case of the aerial cargo delivery the state space is the joint position-velocity vector of the quadrotor's and load's centers of the masses, $\x = [\x_p\, \x_v \, \myvec{\eta} \,\dot{\myvec{\eta}}] \in \R{10},$ 
where $\x_p=[x\,y\,z]^T$ is the center of the mass of the UAV, its linear velocities are $\x_v=[\dot{x}\,\dot{y}\,\dot{z}]^T$, and the angular  position $\myvec{\eta} = [ \psi\,\phi]^T$ of the suspended load in the spherical coordinates system originating at the quadrotor's center of mass, and its angular velocities $\dot{\myvec{\eta}} = [ \dot{\psi}\,\dot{\phi}]^T$. 

The action space, $\U \subset \R{d_a}$ is the space of all possible actions that the robot can perform. For the indoor navigation the action space is a two-dimensional vector of wheel speeds, $\uv = (v_l, v_r) \in \R{2},$ while for the aerial cargo delivery the action space is an acceleration vector applied to the quadrotor's center of the mass $\uv \in \R{3}.$ .

The transition probability, $P:\X \times \U \rightarrow \R{}$ is a probability distribution over state and actions.
Like many reinforcement learning systems, we assume a presence of a simplified black-box simulator without knowing the full non-linear system dynamics.
The simulator for the indoor navigation task is a kinematics simulator operating at 5~Hz. To simulate imperfect real-world sensing, the simulator adds Gaussian noise, $\gauss{0}{0.1}$, to its observations. The simulator for the aerial cargo delivery task operates at 50~Hz because of the inherent instability of the quadrotor, and is a simplified model of the quadrotor-load model described in \cite{faust-ai-journal}. 

During training the agent observes a scalar reward, $R:\X \rightarrow \R{},$ and learns a policy $\pi(\x) = \uv,$ that, given an observed state $\x \in \X$, returns an action $\uv \in \U$ that the agent should perform to maximize long-term return, or value,
\begin{equation}
\label{eq:greedy}
\pi^*(\x) = \argmax_{\ac \in \U} E\left(\sum_i \gamma^i R(\x_i)\right).
\end{equation}
We reward the agent for reaching the goal, with task specific reward shaping terms for each of the two tasks. For the indoor navigation task, we reward the agent for staying away from obstacles, while for the aerial cargo delivery task we reward minimizing the load displacement. 

We train the agent with a continuous action RL algorithm in a small environment. We choose continuous action RL because, although more difficult to train, they provide more precise control \cite{faust-acta-13, ddpg}, and faster policy evaluation time \cite{faust-acta-13}. We train indoor navigation tasks with DDPG \cite{ddpg} in a small space (Fig. \ref{fig:mtv1965}). The aerial cargo delivery agent is trained with CAFVI \cite{faust-acta-13} using only \dist{1} around to goal for the training.
Trained once, the agents can be used in different environments to plan a trajectory by observing the world to receive an observation or state, evaluating the policy \eqref{eq:greedy}, and applying the recommended action to the state. Note that agent plans in the state space, which is generally higher dimensionality than the C-space.

\subsection{PRM construction}
\label{ss:mp}

The basic PRM method works by performing a uniform random sampling of robot configurations in the the robot's configuration space, retaining only collision-free samples as nodes in the roadmap. PRMs then attempt to connect the samples to their nearest neighbors using a local planner. If there is an obstacle-free path between two nodes, the edge is added to the roadmap.

We modify the basic PRM by changing the way nodes are connected. Since, we are primarily interested in robustness to noise and adherence to the task, we only connect two configurations if the RL agent can consistently perform the point-to-point task between the two points. Because the state space $\X$, is a superset of the C-space, we sample multiple variations around the start and goal configuration points, and add an edge only if the success-rate exceeds a threshold. Note that this means that PRM trajectories cannot guarantee to be collision-free paths. That is not a limitation of the method, but rather the results of sensor noise.
Nevertheless, later when discussing the results, we estimate a lower bound on the probability of collision.

Algorithm \ref{alg:add_edge} describes how PRM-RL adds edges to the PRMs. We sample multiple points from the state space, which correspond to the start and goal in the configuration space, and attempt to connect the two points. An attempt is successful only if the agent reaches sufficiently close to the goal point. 
To compute the total length of a trajectory, we sum the distances for all steps plus the remaining distance to the goal. The length we associate with the edge is the average of the distance of successful edges. The algorithm recommends adding the edge to the roadmap if the success rate is above a predetermined threshold.  If too many unsuccessful trials are attempted, the method terminates. The number of collision checks in Algorithm \ref{alg:add_edge} is $O(max_{steps} * num_{attempts})$, because there are multiple attempts for each edge. Each trial of checking the trajectory can be parallelized with $num_{attempts}$ processors.

\begin{algorithm}[h!b]
	\caption{PRM-RL Add edge} 
	\label{alg:add_edge}
	\begin{algorithmic}[1]
		\INPUT $s,\,g \in C_{space}$: Start and goal.
		\INPUT $p_{success} \in [0, 1]$ Success threshold.
		\INPUT $num_{attempts}$: Number of attempts.
		\INPUT $\epsilon$: Sufficient distance to the goal.
		\INPUT $max_{steps}$: Maximum steps for trajectory.
		\INPUT $L(\x)$: Task predicate.
		\INPUT $\pi$: RL agent's policy.
		\INPUT $D$ Generative model of system dynamics.
		\OUTPUT $add_{edge}, success_{rate}$ , $length$
		\STATE $success \leftarrow 0$, $length \leftarrow 0$
		\STATE $needed \leftarrow p_{success} * num_{attempts}$
		\FOR {$i=1,\cdots num_{attempts}$ /* Run in parallel.*/}
		\STATE $\x_s \leftarrow s.SampleStateSpace()$ // Sample from the
		\STATE $\x_g \leftarrow g.SampleStateSpace()$ // state space
		\STATE $success_{rate} \leftarrow 0$, $steps \gets 0$, $\x \gets \x_s$
		\STATE $length_{trial} \leftarrow 0$
		\WHILE {$L(\x) \wedge steps < max_{steps} \wedge 
			\|p(\x)-p(\x_g)\| > \epsilon \wedge p(\x) \in$ C-free  }
		\STATE $\x_p \leftarrow \x$, $\uv \leftarrow \pi(\x)$
		\STATE $\x \leftarrow D.predictState(\x, \uv)$
		\STATE $num_{steps} \leftarrow num_{steps} + 1$
		\STATE $length_{trial} \leftarrow length_{trial} + \|\x - \x_p \|$
		\ENDWHILE
		\IF {$\|p(\x)-p(\x_g)\| < \epsilon$}
		\STATE $success \leftarrow success + 1$
		\ENDIF
		\IF {$needed > success \wedge i > needed$}
		\RETURN False, 0, 0 // Not enough success, we can terminate.
		\ENDIF
		\STATE $length_{trial} \leftarrow length_{trial} + \| p(\x) - p(\g) \|$
		\STATE $length \leftarrow length + length_{trial}$
		\ENDFOR
		\STATE $length \leftarrow \frac{length}{success}$, $success_{rate} \leftarrow \frac{success}{i}$
		\RETURN  $success_{rate} > p_{success}, success_{rate}, length$
	\end{algorithmic}
\end{algorithm}

\subsection{PRM-RL Querying}
To generate long-range trajectories, we query a roadmap, which returns a list of waypoints. A higher-level planner then invokes a RL agent to produce a trajectory to the next waypoint. When the robot is within the waypoint's goal range, the higher-level planner changes the goal with the next waypoint in the list. 

\section{Results}
\label{sec:4}
In this Section we evaluate the performance of PRM-RL for the indoor navigation and aerial cargo delivery tasks.

\subsection{Indoor Navigation}
\label{sec:res:indoor}

We work with four maps depicted in Fig. \ref{fig:maps}. We train the RL agent to avoid the obstacles in a \dist{14} by \dist{17} environment (Fig. \ref{fig:mtv1965}) using the DDPG algorithm; our setup follows \cite{ddpg} but with two hidden layers in our actor networks (34, 55), two joint hidden layers in the critic network (163, 33) with an additional hidden layer of (261) for states and a weight decay of 0.01, trained with batch size 124, with the Adam optimizer with alpha 0.9, beta 0.999, epsilon 1e-08, learning rate 7.37e-05 for the actor and 1.14e-04 for the critic, with a target network updated every 13 training steps, and with a replay buffer with 200K entries. When training and using the RL agent, the goal tolerance is \dist{0.5}. This enables the RL agent to train in the presence of noisy sensors and dynamics and is necessary because the agent does not rely on external localization.  Recall that the only input to the agent is position of the goal, and the noisy LIDAR data (Fig. \ref{fig:lidar}). 
The evaluation environments (Fig. \ref{fig:b40}, \ref{fig:sfo}, and \ref{fig:1667}) are between 12 and 52 times larger than the training one. 

We evaluate the PRM-RL by comparing them with PRMs built with a straight line planner (PRM-SL). We do not compare with RRTs because they are one-time planners and are prohibitively expensive for building on-the-fly.
Each roadmap is evaluated on 100 queries selected from the C-free space.  We examine 1) the cost of building the roadmaps; 2) the qualities of the planner trajectories; 3) the actual performance of the agent in simulation; and 4) the experimental results.

\subsubsection{Roadmap construction evaluation}
To build the PRMs, we use an $85\%$ success rate to connect the edges, over 20 trials. The PRMs attempts to connect all the nearest neighbors with within \dist{10} from a node. We construct roadmaps for three different node densities: $0.1$, $0.2$, and $0.4$ samples per meter squared.

Table \ref{tab:roadmap_stats} summarizes the roadmap characteristics. We examine the number of nodes in the roadmap, number of edges, and collision checks performed to build the roadmap, and include percent of success for 100 randomly generated queries. As expected, across all environments and roadmap construction methods, the higher sampling density produces larger maps and more successful queries. The number of nodes in the map does not depend on the local planner, but the number of edges and collision checks do. The number of collision checks is approximately between 10 and 20 times higher for the RL local planner using  Algorithm \ref{alg:add_edge}, because we terminate collision checks early for the edges that consistently fail.  The SL planner does not use noisy sensor observations, and therefore, requires a single trial to add or reject an edge.  We observe that roadmaps built with the RL local planner are more densely connected with $15\%$ and $50\%$ more edges. This is expected because the RL agent is able to go around the corners and small obstacles where the straight line planner cannot.

\begin{table*}
	\caption{Roadmap construction summary different node sampling densities ($0.1$, $0.2$, and $0.4$ samples per meter squared). Environment, method, success rate on 100 query evaluation, number of nodes, number of edges, number of collision checks.}
	\label{tab:roadmap_stats}
	\centering
	\begin{tabular}{l|l|rrr|rrr|rrr|rrr}
		Sampling    & Method  & \multicolumn{3}{|c}{Query success rate (\%)}  & \multicolumn{3}{|c}{Nodes}  & \multicolumn{3}{|c}{Edges}  & \multicolumn{3}{|c}{Collision Checks} \\ 
		density & & 0.1	& 0.2	& 0.4  &	0.1	& 0.2	& 0.4  &	0.1	& 0.2	& 0.4  &	0.1	& 0.2	& 0.4\\ \hline
		Training &  PRM-RL &  0.32 &  0.36 &  0.50 &  16 &  32 &  63 &  33 &  166 &  663 &  13009 &  38292 &  223898 \\
		&  PRM-SL &  0.06 &  0.09 &  0.29 &  16 &  32 &  63 &  15 &  123 &  464 &  914 &  3649 &  17264 \\ \hline
		Building 1 &  PRM-RL &  0.14 &  0.31 &  0.43 &  436 &  871 &  1741 &  3910 &  15632 &  59856 &  1476931 &  5755744 &  23303949 \\
		&  PRM-SL &  0.06 &  0.17 &  0.15 &  436 &  871 &  1741 &  3559 &  13937 &  52859 &  156641 &  622257 &  2393841 \\ \hline
		Building 2 &  PRM-RL &  0.17 &  0.22 &  0.38 &  116 &  232 &  463 &  403 &  1602 &  6833 &  294942 &  1174655 &  5218619 \\
		&  PRM-SL &  0.06 &  0.11 &  0.18 &  116 &  232 &  463 &  276 &  1190 &  5365 &  18297 &  72850 &  312859 \\ \hline
		Building 3 &  PRM-RL &  0.19 &  0.39 &  0.56 &  441 &  881 &  1761 &  2962 &  11850 &  45623 &  1152524 &  4492144 &  17947728 \\
		&  PRM-SL &  0.07 &  0.11 &  0.08 &  441 &  881 &  1761 &  1852 &  7570 &  30267 &  97088 &  375304 &  1493816 \\ \hline
		
	\end{tabular}
\end{table*}

\subsubsection{Expected trajectory characteristics}
Now we look at the expected performance of PRM-RL across the four environments with respect to 100 randomly selected queries. We focus only on the roadmaps with 0.4 samples per meters squared density. Table \ref{tab:query_expected} summaries the expected number of waypoints, trajectory length, and duration. The SL local planner is more optimistic, expecting 100\% success on the planner trajectories. The RL agent computes the expected probability of success as a joint probability of each edge, since each edge success is an independent random event. Thus, the lower bound on the expected success rate over multiple waypoints is $0.85^{n_w},$ where $n_w$ is number of waypoints. So, the lower bounds on trajectory success for Building 2 for example should be $0.85^{6.05} = 37 \%$, while the lower bound for Building 3 is  $0.85^{12.65} = 13 \%.$ Given that the expected success rates in Table \ref{tab:query_expected} are above 90\%, that means that most of the edges added to the roadmap had 95-100\% success rate during the roadmap construction time. The actual success rates for the RL local planner are between the lower bounds. This RL agent does not require the robot to come to rest at the goal region, therefore the robot experiences some inertia when the waypoint is switched. This causes some of the failures. That said, the actual success rates for the PRM-SL are significantly lower. What's more so, the PRM-SL planner has no estimate of a path risk. We also see that the PRM-RL paths contain more waypoints, with the exception of Building 3. Building 3 is the largest, and the paths through it require more turns. The RL agent can execute some of the these turns without adding a waypoint. Expected trajectory length and duration are longer for the RL agent, because the agent uses more realistic estimates of what the robot can achieve.

\begin{table*}
	\caption{Expected path and trajectory characteristics over 100 queries. Environment, method, actual and expected success percent, number of waypoints in the path, expected trajectory length in meters, and duration in seconds.}
	\label{tab:query_expected}
	\centering
	\begin{tabular}{l|l|r|r|rr|rr|rr}
		Environment & Method & \multicolumn{2}{|c}{Success (\%)} & \multicolumn{2}{|c}{Number of} & \multicolumn{2}{|c}{Trajectory} & \multicolumn{2}{|c}{Duration (s)} \\
		&  & Actual & Expected & \multicolumn{2}{|c}{waypoints} & \multicolumn{2}{|c}{length (m)} & \multicolumn{2}{|c}{} \\\hline
		&&&& $\mu$ & $ \sigma$  & $\mu$ & $ \sigma$ & $\mu$ & $ \sigma$\\ \hline
		Training &  PRM-RL &  50 &  90 & 7.11 & 2.88 & 18.68 & 11.80 & 36.28 & 36.28 \\
		&  PRM-SL &  28 &  100 & 5.44 & 3.51 & 9.39 & 9.69 & 8.29 & 8.29 \\  \hline
		Building 1 &  PRM-RL &  43 &  91 & 12.09 & 10.68 & 56.88 & 63.37 & 107.78 & 107.78 \\
		&  PRM-SL &  15 &  100 & 11.99 & 8.88 & 46.69 & 43.85 & 43.07 & 43.07 \\ \hline
		Building 2 &  PRM-RL &  38 &  95 & 6.05 & 4.46 & 21.54 & 25.82 & 41.69 & 41.69 \\
		&  PRM-SL &  18 &  100 & 6.98 & 5.69 & 18.75 & 23.05 & 16.97 & 16.97 \\ \hline
		Building 3 &  PRM-RL &  56 &  92 & 12.62 & 5.12 & 64.94 & 33.96 & 122.31 & 122.31 \\
		&  PRM-SL &  8 &  100 & 15.58 & 8.02 & 59.03 & 35.47 & 54.00 & 54.00 \\        \hline      
	\end{tabular}
\end{table*}
\excise{
	\begin{table}
		\caption{Expected path and trajectory characteristics over 100 queries. Environment, method, actual and expected success percent, number of waypoints in the path, expected trajectory length in meters, and duration in seconds.}
		\label{tab:query_expected}
		\centering
		\begin{tabular}{l|l|r|r|rr|rr|rr}
			Env. & Method & \multicolumn{2}{|c}{Success (\%)} & \multicolumn{2}{|c}{Num. of} & \multicolumn{2}{|c}{Traj.} & \multicolumn{2}{|c}{Dur. (s)} \\
			&  & Act. & Expt. & \multicolumn{2}{|c}{waypoints} & \multicolumn{2}{|c}{length (m)} & \multicolumn{2}{|c}{} \\\hline
			&&&& $\mu$ & $ \sigma$  & $\mu$ & $ \sigma$ & $\mu$ & $ \sigma$\\ \hline
			Train.&  PRM-RL &  50 &  90 & 7.11 & 2.88 & 18.68 & 11.80 & 36.28 & 36.28 \\
			&  PRM-SL &  28 &  100 & 5.44 & 3.51 & 9.39 & 9.69 & 8.29 & 8.29 \\  \hline
			Bldg.&  PRM-RL &  43 &  91 & 12.09 & 10.68 & 56.88 & 63.37 & 107.78 & 107.78 \\
			1&  PRM-SL &  15 &  100 & 11.99 & 8.88 & 46.69 & 43.85 & 43.07 & 43.07 \\ \hline
			Bldg.&  PRM-RL &  38 &  95 & 6.05 & 4.46 & 21.54 & 25.82 & 41.69 & 41.69 \\
			2&  PRM-SL &  18 &  100 & 6.98 & 5.69 & 18.75 & 23.05 & 16.97 & 16.97 \\ \hline
			Bldg. &  PRM-RL &  56 &  92 & 12.62 & 5.12 & 64.94 & 33.96 & 122.31 & 122.31 \\
			3&  PRM-SL &  8 &  100 & 15.58 & 8.02 & 59.03 & 35.47 & 54.00 & 54.00 \\        \hline      
		\end{tabular}
	\end{table}
}
\subsubsection{Actual trajectory characteristics}
To evaluate the actual indoor navigation task performance, we look at the query characteristics for successful versus unsuccessful queries. 
Table \ref{tab:actual} summarizes the differences in number of waypoints, trajectory length, and duration of the successful and unsuccessful trajectories. The RL agent produces higher success rate than the straight line planner. PRM-RL performs the best in the Building 3, which is the largest, while PRM-SL performs the worst in that environment. The longest successfully executed trajectory is \dist{216} meters long, taking 400 seconds to complete, and passing through 45 waypoints (see accompanying video \cite{prm-rl-video} ).

The successful trajectories have fewer waypoints than the expected waypoints from Table \ref{tab:query_expected}, which means that the shorter queries are more likely to succeed, as is expected. We also see that the unsuccessful queries fail after only few waypoints. The PRM-RL has higher success rate overall and performs the best in the largest environment (Building 3), while the PRM-SL performs the worst in it. Fig. \ref{fig:maps} depicts some of the PRM-RL trajectories the test environments.  

\begin{table*}
	\caption{Characteristics of the successful and unsuccessful trajectories.  Environment, method, actual and expected success percent, number of waypoints in the path, expected trajectory length in meters, and duration in seconds.}
	\label{tab:actual}
	\centering
	\begin{tabular}{l|l|r|rr|rr|rr|rr|rr|rr}
		Environment & Method & Success  & \multicolumn{4}{|c}{Number of waypoints} & \multicolumn{4}{|c}{Trajectory length (m)} & \multicolumn{4}{|c}{Duration (s)} \\
		&  & (\%) &                             \multicolumn{2}{|c}{Successful}  & \multicolumn{2}{|c}{All} & \multicolumn{2}{|c}{Successful}  & \multicolumn{2}{|c}{All} & \multicolumn{2}{|c}{Successful}  & \multicolumn{2}{|c}{All} \\\hline
		&&& $\mu$ & $ \sigma$  & $\mu$ & $ \sigma$ & $\mu$ & $ \sigma$ & $\mu$ & $ \sigma$& $\mu$ & $ \sigma$& $\mu$ & $ \sigma$\\ \hline
		Training &  PRM-RL &  50 & 4.29 & 2.38 & 4.32 & 2.88 & 13.88 & 8.62 & 14.54 & 9.46 & 36.74 & 26.06 & 37.04 & 37.04 \\
		&  PRM-SL &  28 & 2.45 & 3.02 & 2.49 & 2.73 & 6.71 & 7.10 & 6.39 & 6.32 & 6.71 & 7.10 & 6.39 & 6.39 \\\hline
		Building 1 &  PRM-RL &  43 & 10.76 & 11.01 & 8.44 & 9.77 & 51.17 & 56.66 & 43.16 & 51.74 & 117.46 & 108.60 & 112.29 & 112.29 \\
		&  PRM-SL &  15 & 4.37 & 4.69 & 4.09 & 4.51 & 20.19 & 21.78 & 18.32 & 20.46 & 20.19 & 21.78 & 18.32 & 18.32 \\\hline
		Building 2 &  PRM-RL &  38 & 8.08 & 3.67 & 3.96 & 4.53 & 32.93 & 21.58 & 23.64 & 20.78 & 70.18 & 45.87 & 77.33 & 77.33 \\
		&  PRM-SL &  18 & 3.89 & 5.63 & 3.41 & 4.71 & 15.31 & 20.79 & 12.29 & 17.20 & 15.31 & 20.79 & 12.29 & 12.29 \\\hline
		Building 3 &  PRM-RL &  56 & 9.74 & 5.60 & 9.61 & 5.79 & 58.71 & 32.09 & 57.60 & 34.78 & 130.79 & 63.06 & 130.06 & 130.06 \\
		&  PRM-SL &  8 & 5.66 & 5.33 & 5.21 & 4.46 & 22.46 & 19.67 & 21.30 & 17.50 & 22.46 & 19.67 & 21.30 & 21.30 \\\hline      
	\end{tabular}
\end{table*}


\subsubsection{Physical robot experiments}
\begin{figure}[]
	\centering
	\includegraphics[width=0.45\textwidth]{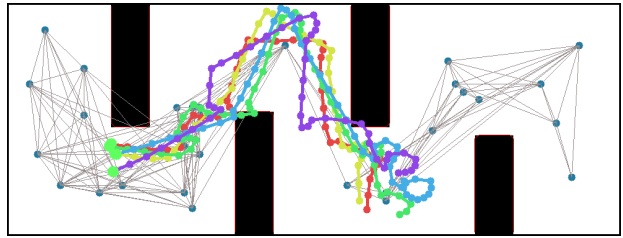}
	\caption{\small Trajectories for a single query, executed on the real differential drive robot over five trails (green, blue, red, purple, and yellow). The grey straight lines indicate graph edge connectivity of the PRM-RL. The trajectories are captured with a mocap system.\label{fig:real_robot}}
\end{figure}

To test the effectiveness and transfer of our approach on a real robot, we created a simple slalom-like environment with four obstacles distributed over an \dist{8} by \dist{3} space.  Fig. \ref{fig:real_robot} illustrates a single query variance due to sensor noise. Each of the trial trajectories reached within the goal region of \dist{0.5} with a mean distance of \dist{0.37}.


\subsection{Aerial Cargo Delivery}
A model of a city, our simulation environment, allows us to test PRM-RL over longer distances, while experimental environments allow us verify the algorithm on a UAV quadrotor equipped with a suspended load. The city model (Fig. \ref{fig:city}), is \dist{450} by \dist{700}, with \dist{200} height. Its bounding box is 250 million times larger than the quadotor's bounding box, and 63 million larger than the RL agent training environment. 
Requiring the load displacement to be under $45^{\circ}$, the evaluation trajectories are result of 100 queries between a fixed origin (depot), and randomly selected goal. We measure load displacement, and trajectory duration, and compare to PRM-SL. The aerial cargo delivery tasks is deterministic, and we build roadmaps requiring one sampling trail, and 100\% success rate. Additionally, this RL agent requires the robot to come at rest at waypoints, resulting in no failures due to inertia.


\subsubsection{Simulation results}
\begin{figure*}[htb!]
	\centering
	
	\begin{tabular}{lll}
		\subfloat[City]{\includegraphics[width=0.3\textwidth, height=40mm]{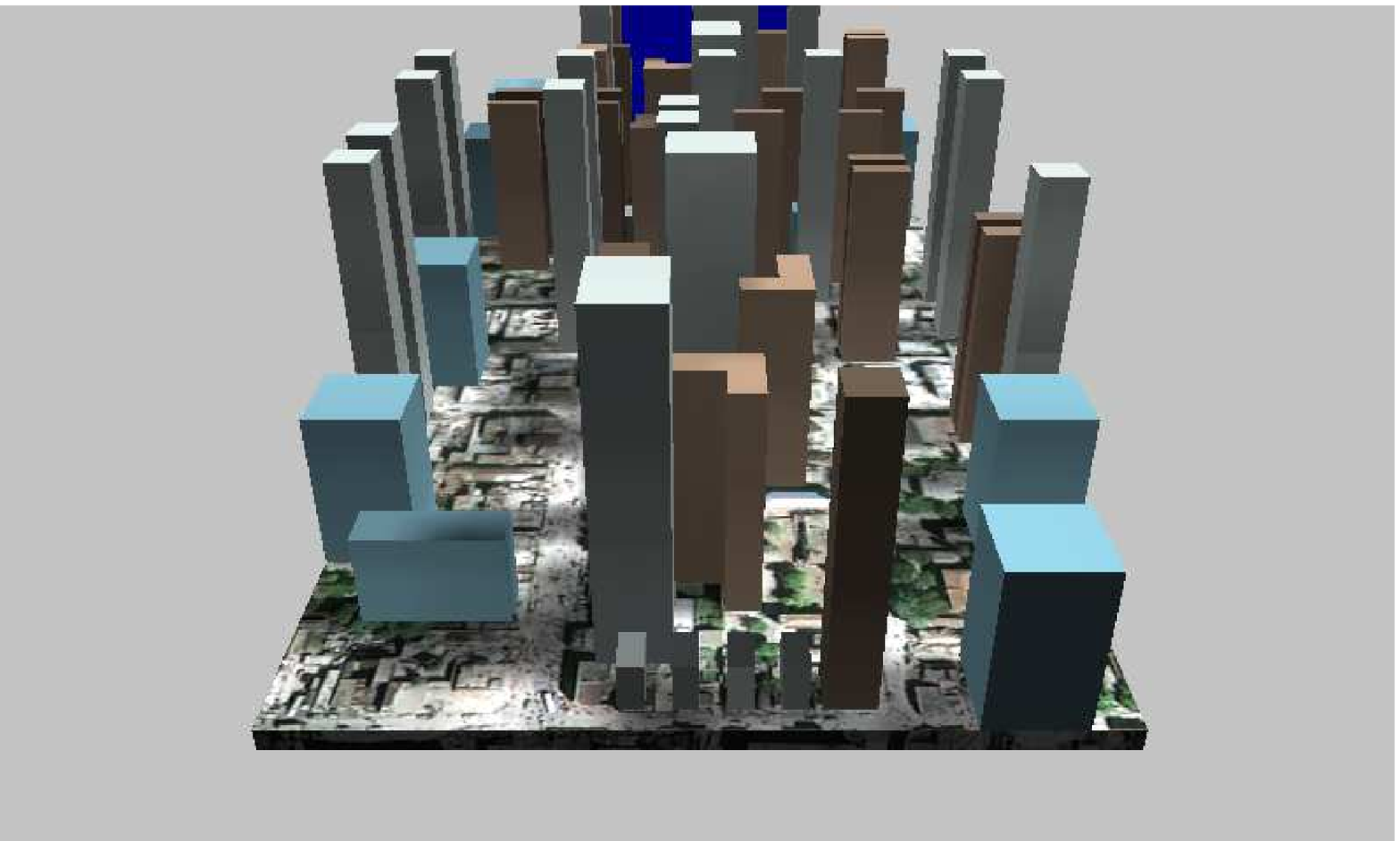}\label{fig:city}} &
		\subfloat[Load displacement]{\includegraphics[width=0.3\textwidth,height=40mm,keepaspectratio=false]{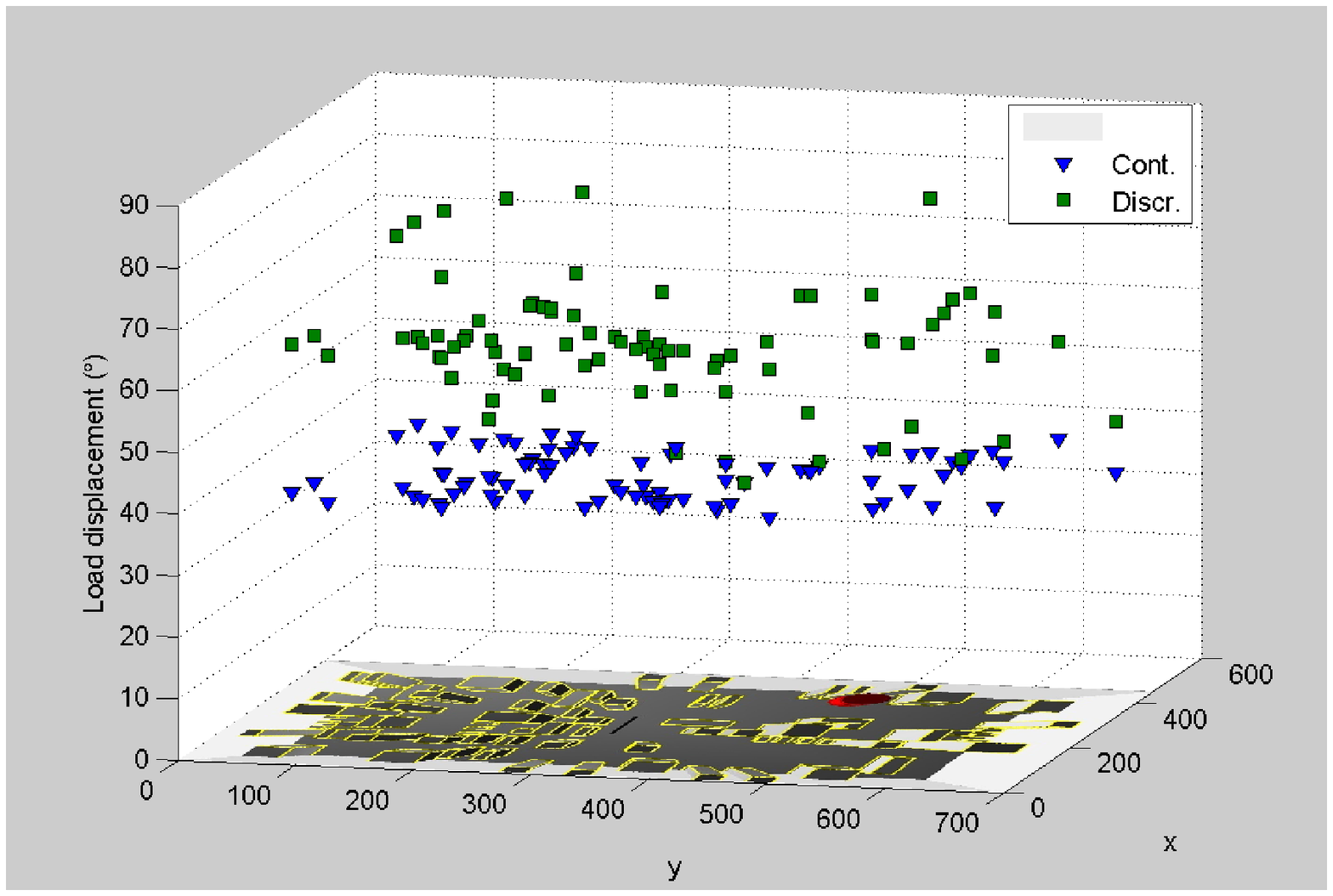}\label{fig:swingXY}} &
		\subfloat[Duration]{\includegraphics[width=0.3\textwidth,height=40mm,keepaspectratio=false]{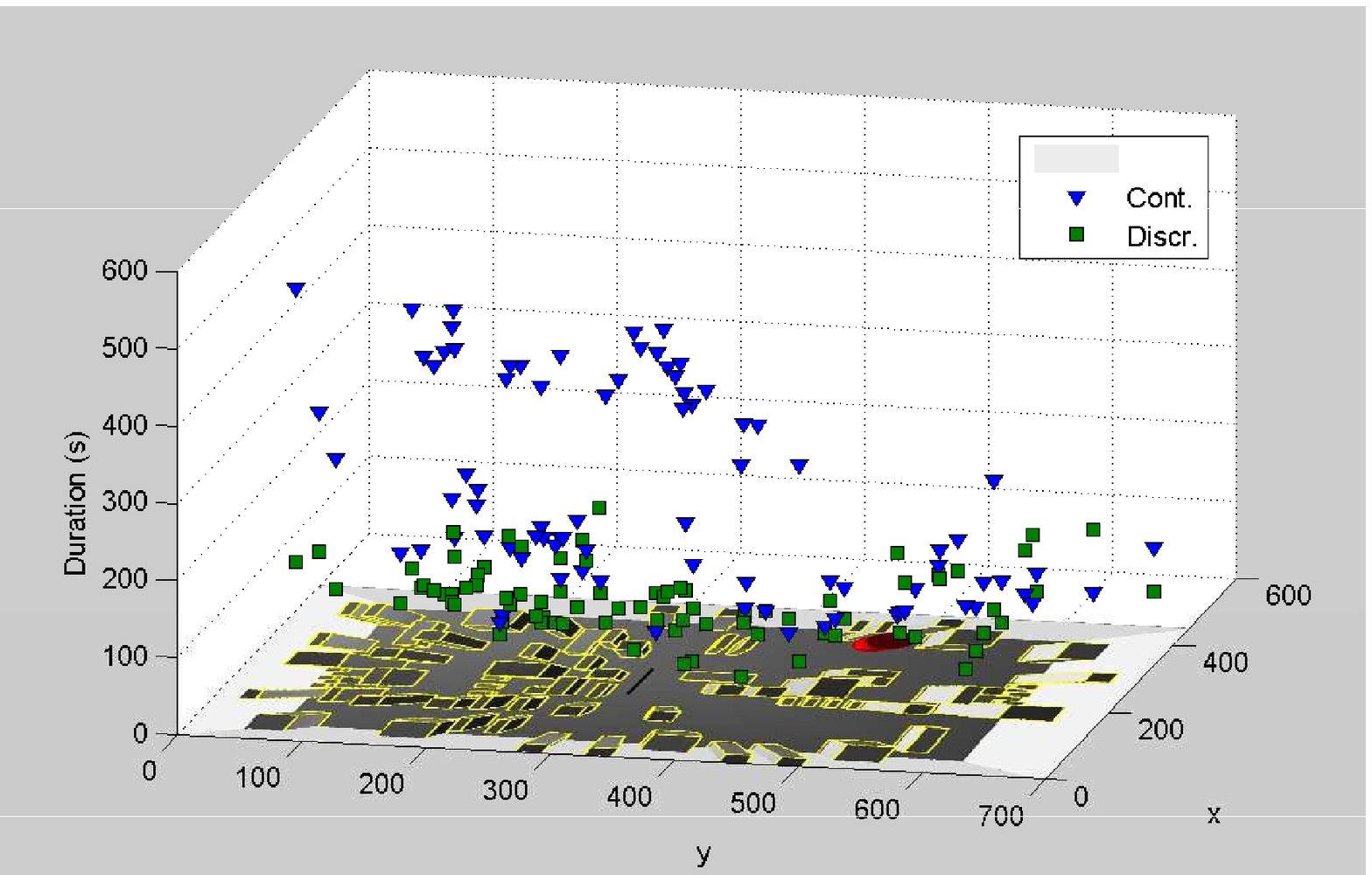}\label{fig:durationXY}}
	\end{tabular}
	\caption{(a) Areal cargo delivery city environment. (b) Load displacement and (c) trajectory duration for PRM-RL (triangle) vs. PRM-SL (square) in the city projected on the $xy$-plane.}
	\label{fig:cityexp}
\end{figure*}

Fig. \ref{fig:cityexp} shows the load displacement and trajectory duration results. The $xy$-plane contains the city projection. The red area marks the origin location. The data points' $x$ and $y$ coordinates are the queries' goal location in the city projected onto $xy$-plane. The data points' $z$-coordinates in Fig. \ref{fig:swingXY} are the maximum load displacement throughout the trajectory, and in Fig. \ref{fig:durationXY} the time it takes to complete the trajectory. Points represented as squares (green) are the result PRM-SL, and the triangle points (blue) are generated with PRM-RL. The longest trajectory executed is over 1 kilometer in length. The maximum load displacement in the continuous case is consistently below the load displacement exhibited with the discrete planner and stays under the required $45^{\circ}.$ Action filtering \cite{faust-ai-journal} guarantees load displacement constrains, but given that it is a discrete action planner with over 1.7 millions action, it adds 1.7 million collision checks per step in the local planner execution. This makes this method computationally prohibitive to build PRMs with. To offset the load displacement control, the PRM-RL trajectories take a longer time to complete the task. They generally contain on average 5 times more waypoints. Fig. \ref{fig:durationXY} shows that in the vicinity of the origin, PRM-RL and PRM-SL trajectories have similar duration. As the goal's distance increases, the discrepancy becomes more significant.

\subsubsection{Experimental results}
\label{ss:exep}
The experiments evaluate PRM-RL on a physical robot to validate the simulation results. We look for discrepancies in the length, duration, and load displacement over the entire trajectory.  The experiments were performed on AscTec Hummingbird quadrotor, carrying a 62-centimeter suspended load weighing 45 grams, in a MARHES aerial testbed \cite{pcf-jurnal-12}. The experimental environments are \dist{3} by \dist{4} by \dist{2}, and contain 3 and 5 obstacles, \dist{2} tall (see video \cite{prm-rl-video}). The quadrotor and load position, tracked via a motion capture system at 100 Hz, require load displacement not to exceed $10^\circ$. 

\begin{figure}[h]
	\begin{minipage}[t]{.5\textwidth}
		\centerline{\includegraphics[width=\textwidth,height=75mm,keepaspectratio=false]{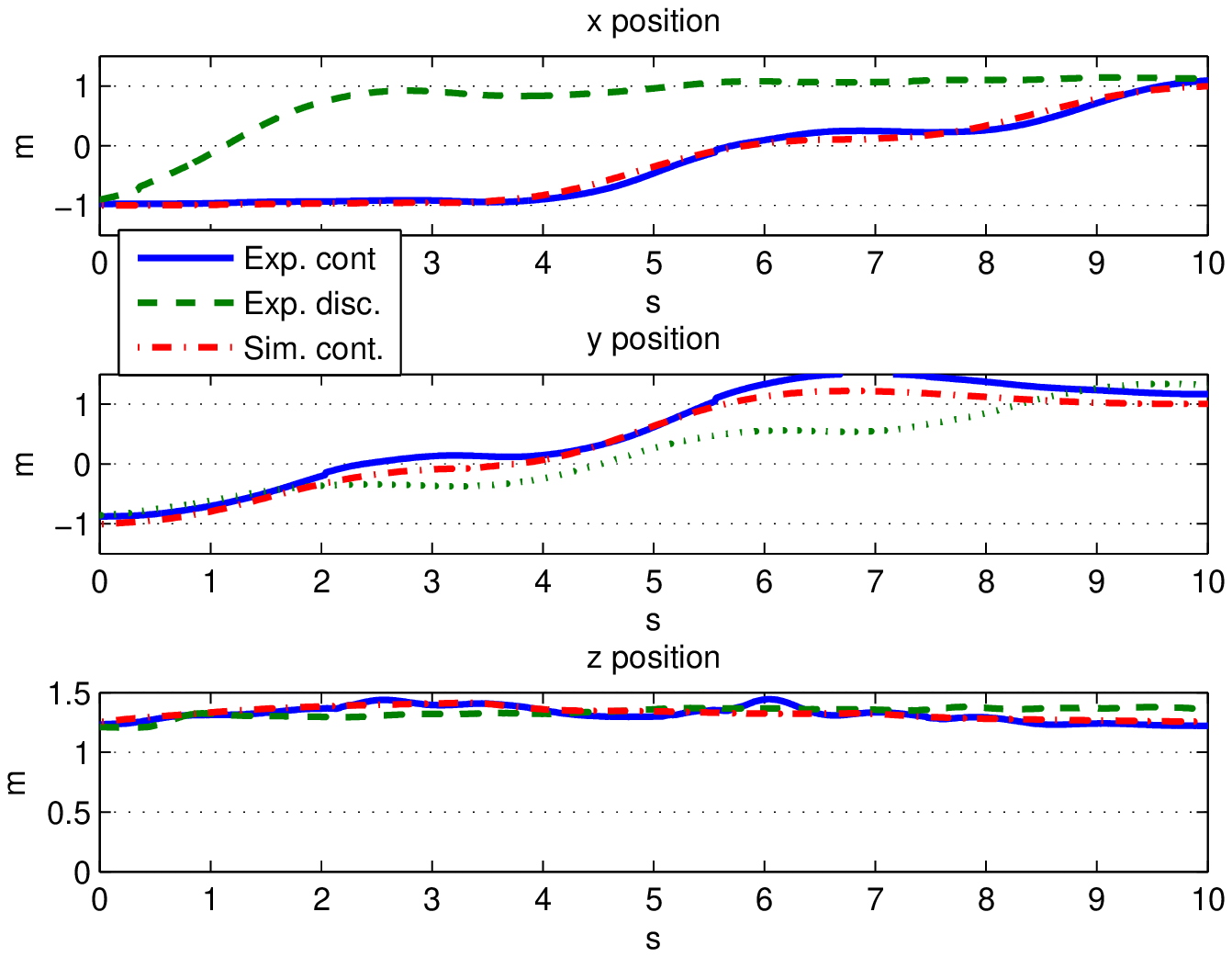}}
		\centerline{(a)}
	\end{minipage}
	\begin{minipage}[t]{.5\textwidth}	
		\centerline{\includegraphics[width=\textwidth,height=45mm,keepaspectratio=false]{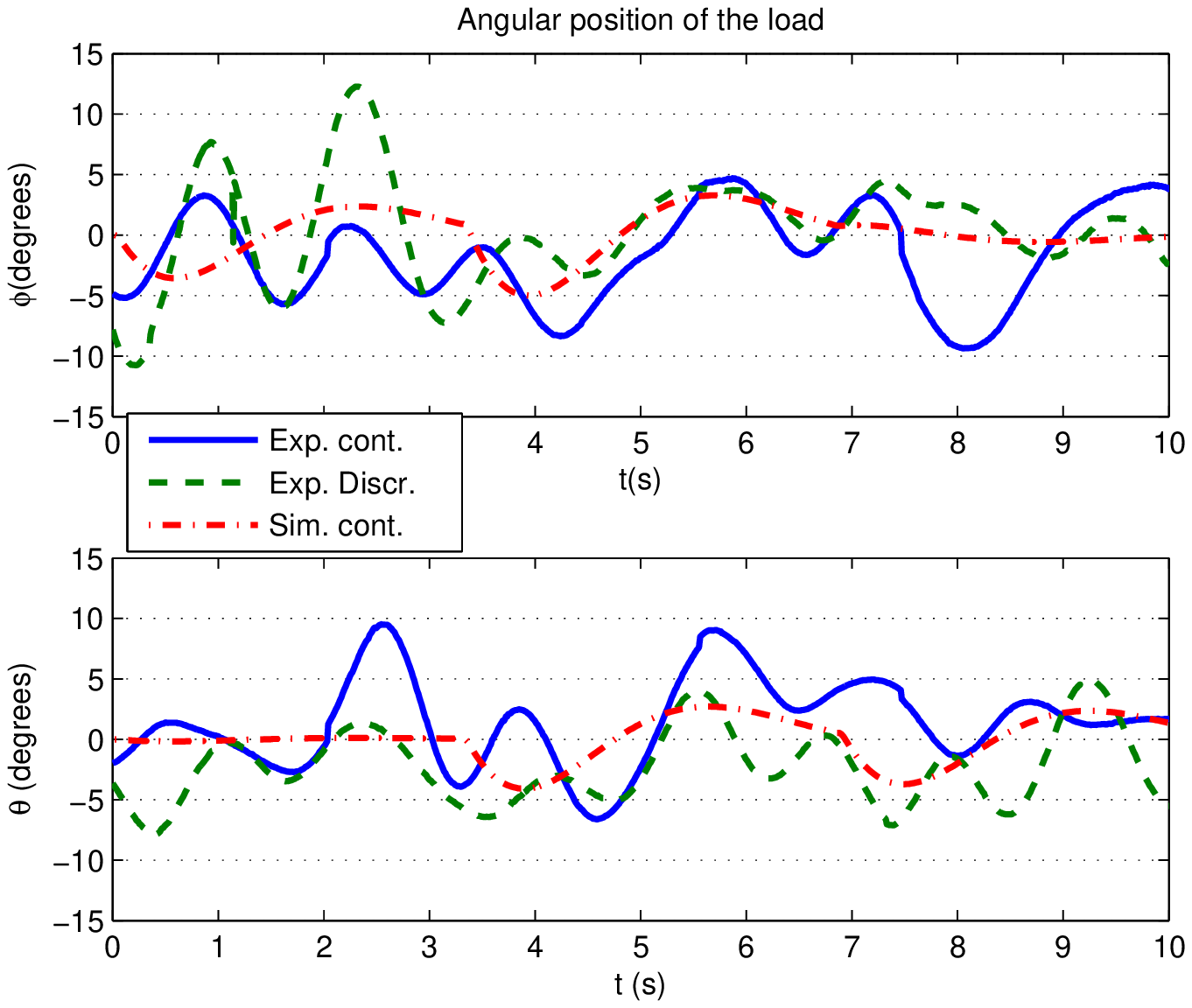}} 
		\centerline{(b)}
	\end{minipage}	
	\caption{Experimental PRM-RL trajectory with three edges, on the quadrotor demonstrating quadrotor (a) and load (b) trajectories.  PRM-RL with continuous action RL from a hardware experiment is shown in comparison to the simulated trajectory.  Also shown for comparison is RL with discrete actions following the PRM-SL path from a hardware experiment.}
	\label{fig:exp}
\end{figure}

Fig. \ref{fig:exp} shows that the vehicle and load trajectories match closely between simulation and experiment, and that the load displacement stays under $10^\circ$, even in experiments. Further, the discrepancy between simulation and experiments of $10^\circ$ remains, due to the unmodelled load turbulence, is consistent with our previous results from \cite{faust-ai-journal}. 

Next, we create a PRM-SL trajectory in the same environment. The vehicle trajectory differs in this case, because the PRMs in the two cases differ. The PRM-SL does not directly control the load displacement, while PRM-RL rejects edges that exceed the maximum allowed load displacement. The load trajectory (Fig. \ref{fig:exp} (b)) indicates that the load displacement exceeds the $10^{\circ}$ limit 2.5 seconds into the flight. The video \cite{prm-rl-video} contains the experiment footage.

\section{Conclusions}
\label{sec:5}
We presented PRM-RL, a hierarchical planning method for long-range navigation tasks, that combines sampling-based path planning with RL agents to complete tasks in very large environments. Evaluated on two case studies, one with noisy sensor feedback task, and the other on a complex unstable dynamics, we showed that PRM-RL expands the capabilities of both RL agents and sampling-based planners. The indoor navigation task successfully completes trajectory over \dist{210} long, and the aerial cargo delivery creates flights over 1 kilometer long, in the planning space 63 million times larger than the agent's training space. Both tasks are verified experimentally on the physical robots.

\section*{Acknowledgements}
\small
We would like to thank Patricio Cruz from MARHES laboratory \cite{marhes} for assisting with experiments, 
Torin Adamson from the Tapia Lab \cite{amprg} for creating virtual environment models, and Parasol lab \cite{parasol} for providing motion planning library. 
Tapia is supported in partby the National Science Foundation under Grant Numbers IIS-1528047 and IIS-1553266.

\bibliographystyle{abbrv}
\bibliography{literature}  

\end{document}